\documentclass{article}

\usepackage[preprint]{neurips_2026}

\usepackage[utf8]{inputenc}
\usepackage[T1]{fontenc}
\usepackage{hyperref}
\usepackage{url}
\usepackage{booktabs}
\usepackage{amsfonts}
\usepackage{amsmath}
\usepackage{amssymb}
\usepackage{array}
\usepackage{graphicx}
\usepackage{nicefrac}
\usepackage{xcolor}
\usepackage{natbib}
\usepackage{multirow}
\usepackage{tikz}
\usetikzlibrary{arrows.meta, shapes.geometric, positioning}
\usepackage{pgfplots}
\pgfplotsset{compat=1.18}

\newcommand{\glossaryterm}[2]{\hyperref[gls:#1]{#2}}

\title{Necessary, Decodable, and Reversible, Yet Not Transferable:\
A Stress Test for Attention-Head Role Claims}

\author{
  \textbf{Philip Quirke} \\
  Martian \\
  \texttt{philip@withmartian.com}
}

\begin{document}

\maketitle


\begin{abstract}
Mechanistic studies often assign a component a role when removing it damages a
behavior, its activation linearly encodes task information, and restoring that
activation repairs the damage. We test the stronger implication: can the same
activation carry the requested computation into another prompt? In the roughly
8 of 15 \glossaryterm{model-family-cell}{model--family cells} receiving our full matched-control transfer assay,
covering three 7--8B instruction-tuned models and five single-step computation
families, tested attention-head states did not produce clean computation
transfer. Same-answer and matched-context controls instead exposed inert or
broad effects. Positive controls bound this result rather than eliminating its
caveats: the instrument recovers known function vectors when writing into an
underspecified prompt, and recovers a compact override-capable residual state in
one of three controlled-model seeds. Neither control proves sensitivity to the
object carried by real-model attention during prompt override. On two composed
forms in Qwen, a strongly decodable operation direction also failed to steer the
generated answer under \glossaryterm{passing-gate}{passing gates}; the corresponding cross-model test was
invalid. These results show that necessity, decodability, same-prompt repair, and
cross-prompt transfer are separable evidence. They do not establish that a
portable selection state is absent from untested components or settings.
\end{abstract}

\begin{figure}[t]
\centering
\resizebox{\textwidth}{!}{%
\begin{tikzpicture}[
  card/.style={rectangle, rounded corners=3pt, draw=black!55, thick,
               text width=5.7cm, minimum height=2.8cm, align=left,
               inner sep=8pt, font=\small},
  arr/.style={->, very thick, black!55, rounded corners=4pt}
]
  \node[card, fill=blue!7] (necessary) at (0,0) {
    \textbf{1. \glossaryterm{necessity}{NECESSITY}}\\[-1pt]
    \emph{Question: Does it matter?}\\[3pt]
    \textbf{Test:} zero the component's output\\
    \textbf{Supports:} measurement of behavior\\
    \textcolor{red!65!black}{\textbf{Doesn't show:} what information it carries}
  };
  \node[card, fill=blue!11] (decodable) at (7.0,0) {
    \textbf{2. \glossaryterm{decodability}{DECODABILITY}}\\[-1pt]
    \emph{Question: Is task information readable?}\\[3pt]
    \textbf{Test:} apply a linear representation audit\\
    \textbf{Supports:} task information is readable\\
    \textcolor{red!65!black}{\textbf{Doesn't show:} model uses the readout}
  };
  \node[card, fill=blue!15] (reversible) at (7.0,-4.2) {
    \textbf{3. \glossaryterm{ablation-reversal}{ABLATION REVERSAL}}\\[-1pt]
    \emph{Question: Can its own state repair damage?}\\[3pt]
    \textbf{Test:} restore clean same-prompt activation\\
    \textbf{Supports:} local repair at the restored token range\\
    \textcolor{red!65!black}{\textbf{Doesn't show:} portable to another prompt}
  };
  \node[card, fill=orange!18] (transfer) at (0,-4.2) {
    \textbf{4. \glossaryterm{cross-prompt-transfer}{CROSS-PROMPT TRANSFER}}\\[-1pt]
    \emph{Question: Can it move the computation?}\\[3pt]
    \textbf{Test:} move the activation under matched controls\\
    \textbf{Supports:} scoped interventional generalizability\\
    \textcolor{red!65!black}{\textbf{Doesn't show:} a role outside the tested site and context}
  };
  \draw[arr] (necessary) -- (decodable);
  \draw[arr] (decodable) -- (reversible);
  \draw[arr] (reversible) -- (transfer);
\end{tikzpicture}
}
\caption{The evidence ladder and the scope of each test.
Our central finding is that passing 1--3 does not imply passing 4.
The four forms of evidence are separable.
}
\label{fig:evidence-ladder}
\end{figure}
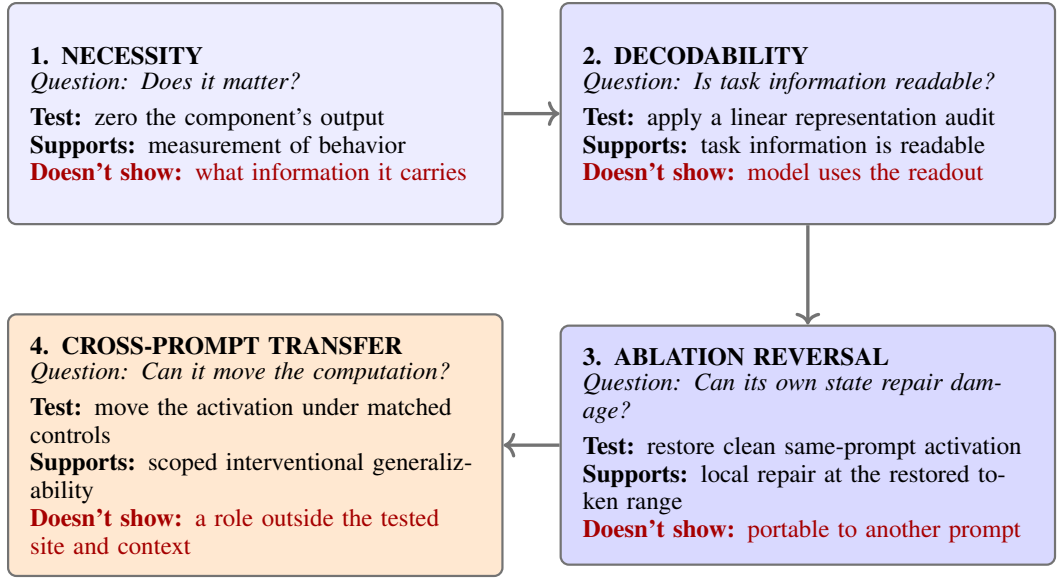


\section{Introduction}

Consider a favorable case for assigning an attention head a computational
role. In Qwen's time-arithmetic prompts, ablating one selected head damages the
target behavior. Its activation contains readable task information, and
restoring that activation at prompt positions repairs most of the ablation
damage. Yet moving the activation from an addition prompt to a subtraction
prompt does not transfer addition. The source-answer score rises, but a
matched same-computation control rises even more. The intervention moved broad
prompt state, not a specific operation.

This example separates four questions that are often collapsed. Does a
component matter for the behavior? Is task information readable from it? Can
its own clean activation repair an ablation in the same prompt? Can its state
change the requested computation in another prompt? We call the answers
\emph{\glossaryterm{necessity}{necessity}},
\emph{\glossaryterm{decodability}{decodability}},
\emph{\glossaryterm{ablation-reversal}{ablation reversal}}, and
\emph{\glossaryterm{cross-prompt-transfer}{cross-prompt transfer}}, respectively.
The last property is also called
\glossaryterm{interventional-generalizability}{interventional generalizability}.
It is stronger than the first three because
it asks whether the proposed content survives a change of context.

Prior work has shown why each weaker test can mislead in isolation. Linear
probes can recover information that a model does not use
\citep{hewitt2019control,elazar2021amnesic}. Patching results depend on the
corruption, site, metric, and window \citep{zhang2024patching}. A successful
subspace edit can exploit a dormant route rather than reveal the mechanism used
on the original input \citep{makelov2023subspaceillusion}. Conversely,
function-vector work shows that some internal states really can be reused in a
new context \citep{todd2024functionvectors}. These observations motivate a
simple stress test: after identifying a behaviorally important and readable
component, try to move its proposed computation under controls that distinguish
semantic transfer from answer or context transfer.

We apply this test to attention heads in three instruction-tuned 7--8B models.
A \glossaryterm{behavioral-screen}{behavioral ablation screen} identifies compact
head sets for arithmetic, comparison, digit, date, and time prompts. A
\glossaryterm{representation-audit}{representation audit} asks what is linearly
readable at those sites. \glossaryterm{same-prompt-restoration}{Same-prompt
restoration} separates components needed during prompt processing from components
supporting answer continuation. Finally,
\glossaryterm{cross-prompt-transfer}{cross-prompt activation transfer} tests whether the source computation
changes the target answer. The full matched-control transfer assay covers
roughly 8 of 15 model--family cells, spanning all three models and all five
families but not every combination.

Our contributions are:

\begin{enumerate}
\item \textbf{An evidence dissociation.} In the tested attention-head cells,
necessity, decodability, and same-prompt repair do not imply clean cross-prompt
computation transfer. We report exact assay coverage rather than treating the
study as an exhaustive model--family grid. The negative also holds across two
Qwen composed forms; an invalid Llama test provides no cross-model evidence.
\item \textbf{Matched controls.}
We distinguish a computation switch from broad movement of source context or
answer-linked state using \glossaryterm{same-answer-control}{same-answer},
\glossaryterm{same-computation-control}{same-computation}, and
\glossaryterm{matched-random-control}{matched-random} controls.
\item \textbf{Sensitivity controls.} The transfer instrument recovers
known function vectors in a \glossaryterm{write-regime}{write regime} and a compact override-capable residual
state in one controlled-model seed. These positives show sensitivity for those
regimes and \glossaryterm{object-class}{object classes}; they do not prove that real-model override failures
are true absences.
\end{enumerate}


\section{Related Work}

We situate this paper within four overlapping waves of mechanistic
interpretability (MI) work: early component-level findings, the common
evidence stack used to support them, the growing catalogue of
\emph{interpretability illusions} that have complicated those findings, and
the recent push toward methodological rigor and auditing.

\paragraph{From induction heads to localized capabilities.}
Early MI work characterized individual attention heads as implementing
identifiable algorithms: induction heads for in-context learning
\citep{olsson2022induction}, name-mover and inhibition heads for indirect
object identification \citep{wang2022ioi}, factual-association retrieval
\citep{meng2022rome}, and copy-suppression heads \citep{mcdougall2023copysuppression}.
A parallel thread showed that capabilities can localize to surprisingly small
parameter or activation subsets: $O(1$--$5)$ attention heads can suffice for
specific measured capabilities \citep{bair2026cscl}, and even a single weight
can be critical for text generation \citep{yu2024superweight}. The
representational counterpart is the linearity hypothesis --- that many
relations and concepts are encoded as linear directions
\citep{hernandez2023linearconcepts} in subspaces shaped by superposition
\citep{elhage2022superposition}.

\paragraph{The common evidence stack.}
These findings rest on a recurring methodological combination: probes for
linear decodability \citep{belinkov2022probing,burns2022ccs},
\glossaryterm{activation-patching}{activation patching} for causal mediation
\citep{vig2020causal,meng2022rome,conmy2023acdc},
and ablation for necessity. The combination is widely employed but has no
community-agreed validation protocol, and probing alone has long been known
to overstate causal claims \citep{elazar2021amnesic,hewitt2019control}.
The patching tool itself has also received scrutiny:
\citet{zhang2024patching} and \citet{heimersheim2024patching} show that within-prompt patching
results are sensitive to corruption method, evaluation metric,
window size, and choice of corrupted tokens. We address
a complementary axis: even when intra-prompt patching follows
current best practices, the resulting evidence does not
establish interventional generalizability, which requires
\emph{cross-prompt} transduction under matched controls.
\citet{hendel2023taskvectors} and \citet{todd2024functionvectors} showed that
some internal states can be extracted and reused as function-like operators ---
a stronger form of evidence that motivates our transfer assays. Importantly,
different head-selection criteria need not identify the same mechanism:
\citet{yin2025attentionheads} distinguish induction from function-vector heads,
and \citet{opielka2026causality} find largely distinct causal and
invariance-encoding head sets.

\paragraph{Interpretability illusions.}
A growing body of work shows that this common combination can produce
confident-looking but misleading conclusions.
\citet{makelov2023subspaceillusion} demonstrate
that subspace activation patching can change a model's output via a
\emph{dormant parallel pathway} causally disconnected from the behavior of
interest, dissociating successful intervention from faithful localization.
\citet{friedman2024illusions} show that simplified proxies (PCA, clustering,
SVD-based summaries) can match the original model in-distribution while
diverging out-of-distribution, undermining the predictive value of derived
mechanistic stories. \citet{meloux2025variance} reframe circuit discovery as
statistical estimation and find that single-input causal mediation scores have
high intrinsic variance, so circuits identified by common pipelines are
fragile under input or hyperparameter perturbations. Our dissociation result
is complementary: rather than questioning a particular tool (subspaces,
proxies, scores), we show that the \emph{joint} evidence stack ---
selective necessity, linear decodability, and ablation reversibility ---
fails to imply transferable computation when probed by cross-prompt transfer
under matched controls.

\paragraph{Toward auditable MI.}
\citet{sharkey2025openproblems} catalogue conflating hypotheses with
conclusions as a recurring failure mode and call for stronger validation
practices. \citet{lan2026auditable} go further, arguing that MI needs a
standardized auditing layer because methodologically inconsistent studies of
the same behavior have already produced conflicting conclusions in the
literature. Our evidence ladder is designed in this spirit: a portable-state
claim is gated on a cross-prompt transfer test under matched controls,
including a \emph{same-answer control} that exposes broad state transfer
masquerading as semantic specificity. The resulting dissociation is consistent
with the illusions literature and operationalizes one path the rigor literature
has called for.


\section{An Evidence Ladder for Role Claims}

The paper asks what each experiment licenses. The tests form an evidence
ladder, but passing one rung does not guarantee the next
(Figure~\ref{fig:evidence-ladder}).

\paragraph{Necessity.}
We first use a behavioral ablation screen to find attention heads whose removal
selectively damages one prompt family. This identifies useful causal targets but
does not tell us why they matter.

\paragraph{Decodability.}
We next ask whether family or operation labels can be read linearly from a
selected head. This is descriptive evidence. Information may be readable at a
site without being used there \citep{hewitt2019control,elazar2021amnesic}.

\paragraph{Same-prompt ablation reversal.}
After ablating a head, we restore its clean activation only at prompt positions
or only at answer positions. This localizes when its contribution is needed. A
\glossaryterm{prompt-side}{prompt-side} repair can still reflect context maintenance
rather than a portable operation state; an
\glossaryterm{answer-side}{answer-side} repair is more naturally interpreted as answer
support in that cell.

\paragraph{Cross-prompt transfer.}
The strongest test moves an activation from a source prompt requiring operation
$A$ into a matched target prompt requiring operation $B$. A portable operation
state should redirect the target toward $A$. Same-computation, same-answer, and
matched-random controls test whether any movement instead follows context,
answer identity, or perturbation magnitude. We require behavioral answer
reordering, not only a small log-probability change.

\subsection{Role Questions, Not Architectural Stages}

It is useful to ask whether a component contributes to recognizing a request,
selecting a computation, or producing an answer. We use these as questions, not
as a claim that transformers implement three serial or disjoint stages. A
component can contribute at several points, and a computation can be distributed
or reconstructed over time. In this paper, a portable computation-selection
state would require prompt-side necessity, decodability, same-prompt repair, and
controlled cross-prompt transfer together. No tested attention head satisfies
all four, but that observation is scoped to the tested interventions.

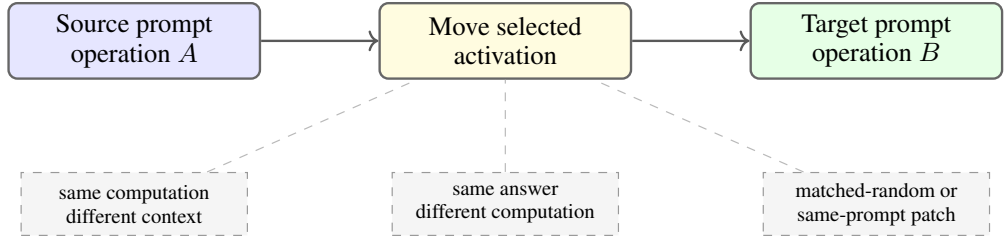
\begin{figure}[t]
\centering
\resizebox{0.95\textwidth}{!}{%
\begin{tikzpicture}[
  node distance=1.1cm and 1.4cm,
  box/.style={rectangle, rounded corners=3pt, draw=black!60, thick,
              minimum width=3.0cm, minimum height=0.9cm, align=center, font=\small},
  ctrl/.style={rectangle, draw=black!40, dashed, fill=gray!8,
               minimum width=2.7cm, minimum height=0.75cm, align=center,
               font=\scriptsize},
  arr/.style={->, thick, black!65}
]
  \node[box, fill=blue!10] (source) {Source prompt\\operation $A$};
  \node[box, fill=yellow!15, right=of source] (patch) {Move selected\\activation};
  \node[box, fill=green!10, right=of patch] (target) {Target prompt\\operation $B$};
  \draw[arr] (source) -- (patch);
  \draw[arr] (patch) -- (target);
  \node[ctrl, below=of source] (samecomp) {same computation\\different context};
  \node[ctrl, below=of patch] (sameans) {same answer\\different computation};
  \node[ctrl, below=of target] (random) {matched-random or\\same-prompt patch};
  \draw[dashed, gray!55] (samecomp) -- (patch);
  \draw[dashed, gray!55] (sameans) -- (patch);
  \draw[dashed, gray!55] (random) -- (patch);
\end{tikzpicture}
}
\caption{Cross-prompt activation transfer and its controls. A computation-specific
effect must redirect the target toward the source operation more than controls
that preserve computation, preserve answer identity, or match perturbation size.}
\label{fig:transfer-design}
\end{figure}


\section{Methods}

\subsection{Models}

We study three 7--8B instruction-tuned models: Qwen2.5-7B-Instruct
\citep{qwen2025qwen25}, Llama-3-8B-Instruct \citep{dubey2024llama3}, and
Mistral-7B-Instruct-v0.2 \citep{jiang2023mistral} (referred to as
\texttt{qwen}, \texttt{llama}, and \texttt{mistral}). The three were chosen as
representative publicly available instruction-tuned models that differ in
architecture and training recipe. Screening and representation analyses cover
all three. The full matched-control transfer assay was run in roughly 8 of the
15 model--family cells formed by the three models and five computation
families. Those cells span every model and family, but not every combination.

\subsection{Prompt Families}

We organize prompt families into three conceptual tiers by presumed circuit complexity:

\begin{itemize}
\item \textbf{Factual recall}: the \texttt{facts} family. Used only as a control
class; these capabilities are expected to rely primarily on lookup rather than
dedicated computation circuits.
\item \textbf{Simple computation}: the primary focus. Families are
\texttt{maths} (arithmetic: addition, subtraction, multiplication, division),
\texttt{compare} (ordered relations: largest, smallest, middle, closest, farthest),
\texttt{digits} (digit properties: odd/even parity, primality, magnitude comparison),
\texttt{dates} (date arithmetic: before/after, day-of-week),
and \texttt{times} (temporal arithmetic: addition/subtraction of clock times).
\item \textbf{Composed computation}: matched Qwen prompts of the form
$(a\mathbin{\pm}b)\times c$ and $(a\mathbin{\pm}b)+c$. These test whether a
decoded inner-operation direction becomes a causal handle when the requested
operation is embedded inside another computation.
\end{itemize}

Simple computation prompt families are hand-built investigative probes motivated by two design principles.
First, anti-correlated subtype pairs (addition vs.\ subtraction, largest vs.\ smallest,
odd vs.\ even) may share representational space while requiring distinct behavior,
making them natural probes for polysemantically packed representations
\citep{elhage2022superposition}. Second, the families span symbolic, numeric, and
temporal domains, providing coverage of diverse computation types at similar complexity.
Each family has subtypes corresponding to distinct requested computations, and the
primary unit of analysis is the \emph{model + prompt-family} combination
(e.g., \texttt{qwen maths}, \texttt{llama digits}). Families are not assumed
to match the model's internal capability ontology; they are experimental levers.

\subsection{Behavioral Ablation Screen}

Following \citet{bair2026cscl}, we zero all outputs of a candidate attention
head across every token position and measure the reference-answer
log-probability change. Selectivity for a head is the target-family damage
minus the maximum damage across non-target families. Heads are ranked by
selectivity score, and we form cumulative top-$k$ sets for $k = 1, 3, 5$ by
greedily adding heads.

We also run two random-mask diagnostics: a main pass using 256
stratified masks (32 heads zeroed per mask) with OMP sparse recovery (20 nonzero
coefficients per family), and a follow-up diagnostic using 1024 masks (8 heads per
mask) to reduce broad-ablation confounds. Subset-lattice evaluation directly measures
every subset of each selected top-5 head set (31 subsets for 5 heads), characterizing
singleton effects, leave-one-out losses, Shapley-style marginal contributions, and
additive residuals.

\subsection{Representation Audit}

We apply \glossaryterm{svd}{SVD (singular value decomposition)}
to head activation matrices to expose interpretable directions
that component-level analysis may miss \citep{ahmad2025svd}. A head may pack
several semantic factors into a shared representational subspace; SVD separates
them.

We collect clean activations for each selected head across a held-out prompt
inventory, then train nearest-centroid probes to decode family and subtype labels from
individual head residual contributions and from selected top-5 concatenated residuals. This
representation audit does not select heads; it asks whether behaviorally important
sites contain linearly decodable requested-computation information.

Critically, we use
\emph{\glossaryterm{counterbalanced-control}{counterbalanced prompt controls}}
to separate semantic
structure from surface artifacts. Pure-surface controls (same answer string, different
computation) and same-format controls are included alongside semantic targets. Dominant
SVD directions that shift under surface controls are flagged as nuisance-dominant.

\subsection{Same-Prompt Restoration}

We zero a selected head set across all token positions, then restore its clean
activations separately at prompt positions and answer positions.
\glossaryterm{recovery-fraction}{Recovery fraction} is the repaired ablation
damage divided by the original ablation
damage. Prompt-side recovery identifies candidates whose contribution is needed
before generation; answer-side recovery identifies components supporting the
scored continuation. This localization does not itself establish what state the
component carries.

\subsection{Cross-Prompt Activation Transfer}

Cross-prompt activation transfer tests whether a head's state can redirect
behavior toward the source computation when patched into a matched target
prompt. We refer to this operation as
\glossaryterm{activation-transduction}{activation transduction} in tables and the
appendix.

We capture clean activations from a
\glossaryterm{source-target}{source prompt} and replace corresponding
\glossaryterm{source-target}{target prompt} activations in selected heads at
selected token positions, then run the
forward pass. Source and target prompts are matched on surface factors (template, item
set, answer format) but differ in the requested computation.

\textbf{Scoring:} We measure reference-answer log-probability change, source-answer
log-probability change, source-vs.-correct margin change, and
\glossaryterm{answer-ordering-change}{answer-ordering changes}
(fraction of prompts where the top candidate changes).

\textbf{Controls.} Three control conditions are essential.
\emph{Same-computation controls} share the requested computation but vary surface
form; if these move as much as computation-changing patches, the effect is not
computation-specific. \emph{Same-answer controls} share the answer string but
differ in computation; if these move comparably, the effect reflects answer-string
familiarity rather than computation-selection transfer.
\emph{\glossaryterm{same-prompt-control}{Same-prompt controls}}
patch the target with its own clean activations; near-zero effect here confirms
that flat results are not attributable to patch instability.

Interventional generalizability requires that computation-changing patches
outperform matched controls and produce answer-ordering changes. Soft logprob
movement without answer reordering shows that a patch affects the model, not
that it transfers a computation.

\subsection{Sensitivity Controls}

We use two positive-control regimes. First, following function-vector work
\citep{todd2024functionvectors}, we fit a mean task direction from in-context
examples and inject it into an underspecified target. This \emph{write} regime
tests whether the same mid-layer additive intervention and candidate-logprob
readout used by our negatives can recover a known reusable state.

Second, we train a two-layer, four-head transformer on balanced mod-10 addition
and subtraction. Across three random seeds, we use cue-position interchange to
determine whether a compact override-capable residual carrier formed. Where it
did, we test whether the mean-difference additive intervention recovers it
against same-cue and norm-matched random controls. This controlled model tests
the transfer operation for a residual object in the override regime. It does
not validate head localization or the representation audit.

\subsection{Composed-Task Injection}

For Qwen, we fit a mean residual direction separating the inner $+$ and $-$
operations in composed prompts. During greedy generation we add the direction
at the layer of strongest held-out separation. The run is valid only if it has
enough baseline-correct items, a prompt-substitution sanity check passes, at
least one dose preserves same-operation answers, and that dose measurably
perturbs output. The primary outcome is the fraction of generations equal to
the source-operation answer, compared with same-operation, mixed-label, and
norm-matched random controls.

\subsection{Statistical Reporting}

For the post-submission residual-direction assays, we report
Clopper--Pearson 95\% intervals, risk differences against matched controls with
bootstrap intervals, Fisher exact tests, and Benjamini--Hochberg correction
within each test family. A zero source-answer rate is treated as a bounded
negative only when the assay's pre-registered validity gates pass. Invalid
cells contribute no positive or negative causal evidence. The earlier
attention-head transfer matrix reports point estimates and matched controls;
formal per-cell interval and multiplicity coverage for that legacy matrix
remains incomplete.


\section{Results}

\subsection{The Behavioral Screen Finds Compact Causal Targets}

Across the three models we screen five computation families plus a factual
control, giving 18 model--family cells (Table~\ref{tab:css-selectivity}). The
screen finds a cleanly selective top-5 head set in 13 cells. Two further cells
damage the target but also have broader collateral effects, and three fail the
selectivity criterion.

Direct subset evaluation confirms that the selected effects are reproducible
under direct ablation and reveals substantial singleton/aggregate variation. Some model + prompt families
are dominated by one head (e.g. \texttt{qwen maths} by \texttt{L23H12}), while others 
are 5-head aggregates (e.g. \texttt{llama digits}). Random-mask recovery
partially supports these findings, recovering the dominant \texttt{qwen maths}
head and the \texttt{llama digits} rank-1 head, but not all top-5 aggregates. 

A separate random-set audit finds that the selected sets are unusual in its
audited cells, although several Llama and Mistral distributions are broad-tailed
rather than sharply sparse. This does not provide formal uncertainty for the
legacy transfer effects. The screen supplies causal targets, not a mechanistic
interpretation.

\subsection{Task Information Is Readable at the Selected Sites}

A representation audit shows that selected heads are not black boxes. Top-5
concatenated residual readouts achieve family-level classification accuracies of
0.77--0.91 in \texttt{qwen}, 0.79--0.93 in \texttt{llama}, and 0.76--0.93 in
\texttt{mistral}. Individual heads can also carry fine-grained subtype
information: \texttt{llama digits} rank-1 (\texttt{L28H19}) reaches subtype
accuracy 0.920, and \texttt{mistral digits} rank-1 (\texttt{L15H0}) reaches 0.840.
SVD decomposition exposes real intra-head structure, consistent with multiple
semantic directions packed into shared representational space. However,
counterbalanced prompt controls reveal that dominant SVD directions are sometimes
nuisance-dominant (surface-sensitive) rather than semantically primary.

These results establish readable structure at behaviorally important sites.
They do not show that the model uses the linear readout or that the selected
head stores a portable operation.

\subsection{Worked Case: Same-Prompt Repair Without Specific Transfer}

The Qwen time-arithmetic cell illustrates the full argument. Ablating the
selected top-5 set causes target-family damage of 1.875 nats with a maximum
control-family damage of 0.486. Its leading head, \texttt{L0H0}, is prompt-side:
restoring its clean prompt activation repairs 0.82 of the ablation damage,
whereas restoring answer positions repairs 0.06.

This is a favorable transfer target, but its source activation is not a clean
operation switch. Moving the head state from a time-addition prompt into a
matched time-subtraction prompt raises the source-answer log-probability by
1.209. A same-computation source raises it by 1.338. Source answers,
operation-related candidates, and format distractors move together. The patch
therefore transports broad prompt context rather than time addition itself.
Necessity, readable task information, and prompt-side repair all survive; the
portable-operation interpretation does not.

\subsection{The Dissociation: Four Properties Come Apart}

Same-prompt restoration localizes whether a selected contribution is needed
during prompt processing or answer production. Table~\ref{tab:dissociation}
shows representative attention-head rows that also received cross-prompt
transfer tests.

\begin{itemize}
\item \textbf{Selected sets are heterogeneous.} Within a single top-$k$ set,
ranks frequently split into answer-side and prompt-side contributions. Some
cells (\texttt{qwen maths}) are dominated by an answer-side head; others
(\texttt{llama digits}) mix prompt-side and answer-side ranks; aggregate top-5
patches therefore conflate distinct roles.

\item \textbf{Prompt-side recovery does not imply cross-prompt transfer.}
Of the eight rows in Table~\ref{tab:dissociation} marked prompt-side ($\checkmark$
in column~3), all eight fail property~(4): source~$\Delta$ is inert, negative,
or contaminated by control movement.

\item \textbf{Same-computation controls expose broad context transfer.} Where
source~$\Delta$ is positive (e.g.\ \texttt{qwen times}), same-computation controls
move equally, so the patch is not computation-specific.
\end{itemize}

Every representative row tested for property~(4) fails clean cross-prompt transfer.
Figure~\ref{fig:scatter} (Appendix~\ref{app:numerical}) shows the prompt-all
vs.\ answer-all clusters; per-head numbers are in Table~\ref{tab:restore}.
Detailed case studies for all model + prompt-family cells, plus the compare-SV2
closest positive, are in Appendix~\ref{app:casestudies}.

\begin{table}[t]
\centering
\caption{Representative evidence matrix for heads and head sets tested with cross-prompt transfer.
(1)~Selective necessity; (2)~linear decodability; (3)~prompt-side ablation reversibility
(prompt-all recovery $\geq 0.7$, or $\times$ if answer-side); (4)~interventional
generalizability (source~$\Delta$ on opposite-computation rows, controlled). These rows are not
an exhaustive model--family grid.}
\label{tab:dissociation}
\setlength{\tabcolsep}{8pt}
\resizebox{\textwidth}{!}{%
\begin{tabular}{lllcccl}
\toprule
Model & Family & Head & (1) & (2) & (3) Pr-all & (4) Src $\Delta$ \\
\midrule
\texttt{qwen} & \texttt{ maths}     & \texttt{L23H12}       & \checkmark & \checkmark
  & $\times$ (0.02) & n/a (answer-side) \\
\texttt{qwen} & \texttt{ times}     & \texttt{L0H0} rank 1  & \checkmark & \checkmark
  & \checkmark (0.82) & $\times$ (broad) \\
\texttt{llama} & \texttt{ digits} & \texttt{L0H5} rank 3  & \checkmark & \checkmark
  & \checkmark (1.00) & $\times$ ($+$0.001) \\
\texttt{llama} & \texttt{ compare} & \texttt{L8H11} rank 2 & \checkmark & \checkmark
  & \checkmark (0.95) & $\times$ ($-$0.21) \\
\texttt{llama} & \texttt{ compare} & \texttt{L6H4} rank 3  & \checkmark & \checkmark
  & \checkmark (1.04) & $\times$ ($-$0.02) \\
\texttt{mistral} & \texttt{ dates} & \texttt{L25H12} rank 1 & \checkmark & \checkmark
  & \checkmark (1.02) & $\times$ ($+$0.001) \\
\texttt{mistral} & \texttt{ dates} & top-5 & \checkmark & \checkmark
  & \checkmark (0.98) & $\times$ ($-$0.006) \\
\texttt{mistral} & \texttt{ compare} & \texttt{L8H1} rank 1 & \checkmark & \checkmark
  & \checkmark (0.95) & $\times$ (broad) \\
\texttt{mistral} & \texttt{ compare} & top-5 & \checkmark & \checkmark
  & \checkmark (0.83) & $\times$ (broad) \\
\bottomrule
\end{tabular}
}
\end{table}

\subsection{Soft Score Movement Is Not a Computation Switch}
\label{sec:casestudies}
\label{sec:comparesv2}

Outside the behavioral screen, an SVD-identified subspace (\texttt{SV2:SV5}) of
Qwen head \texttt{L14H15} carries ordered-selection
information across numbers, digits, letters, dates, and times. Calibrated downstream
injection at layer~18 produces a small but real source-logprob lift ($+0.222$,
scaling monotonically with $\alpha$); however, adversarial low-margin rows show no
answer-ordering change at $\alpha = 0.80$, and same-answer controls move comparably.
This is evidence for a decodable neighborhood and a soft causal bias, not for
operation transfer. In particular, it cannot serve as the positive control for
the negative head-transfer results.

\subsection{Sensitivity Is Demonstrated Only in Bounded Regimes}

Table~\ref{tab:sensitivity} separates what the positive controls establish from
what remains unresolved. In Qwen's write regime, a known function vector raises
the domain-matched compare-task success rate from 0.267 to 0.800 using the same
mid-layer additive construction and candidate readout as the negative assays.
The same vector fails to override an explicit opposite cue (0.00 to 0.00), so
this positive does not validate the override regime.

In the controlled model, all three seeds learn balanced mod-10 addition and
subtraction perfectly. Only seed 1 forms a strong compact cue-position residual
carrier (interchange transfer 0.97). In that seed, the mean-difference
intervention produces a dose-monotone override from 0.04 to 0.94, compared with
0.33 for the strongest norm-matched control. Seeds 0 and 2 do not form a strong
carrier and are inconclusive, not negative sensitivity tests. Thus the result is
an existence proof for one residual object and one seed, not a full-pipeline or
attention-level validation.

\begin{table}[t]
\centering
\caption{Sensitivity evidence and its scope. ``Positive'' means the intervention
recovers a state independently shown to be present; invalid instruments contribute
no evidence.}
\label{tab:sensitivity}
\small
\resizebox{\textwidth}{!}{%
\begin{tabular}{p{2.5cm}p{2.4cm}p{3.2cm}p{4.0cm}}
\toprule
Setting & Object and regime & Result & Licensed conclusion \\
\midrule
Qwen function vector & residual, write & 0.267 $\to$ 0.800 & sensitive in this write configuration \\
Controlled model, seed 1/3 & residual, override & 0.94 versus 0.33 random control & sensitive in principle to a compact residual override state \\
Real-model explicit-cue override & attention-mediated behavior & no matched positive & absence is not licensed \\
\bottomrule
\end{tabular}
}
\end{table}

\subsection{A Qwen-Only Extension to Composed Computations}

The single-step prompts leave open whether a portable operation state appears
only when the model must compose operations. We therefore fit the inner $+$
versus $-$ direction for $(a\mathbin{\pm}b)\times c$ in Qwen. The operation is
strongly decodable, and all validity gates pass, including a dose-efficacy
floor. Yet the generated source-operation answer occurs in 0 of 32 trials in
each direction and no matched control is lower. The Clopper--Pearson 95\% upper
bound is 0.109, below the pre-specified 0.30 reference effect.

The result replicates on $(a\mathbin{\pm}b)+c$: source-answer rate is again
0/32 with a 0.109 upper bound under passing gates. This rules out a large effect
for the tested Qwen direction, site, dose-validity rule, and greedy readout; it
does not rule out item-specific, distributed, or differently situated states.
For the multiplication form, the dose-efficacy floor was met strongly in only
one direction; the reverse direction was nearly inert, so the two directional
nulls are not equally supported. Neither form has a site-matched positive
control proving that a residual intervention at the selected layer can carry a
selection state.
A Llama cross-model attempt failed its harness before injection and is
inferentially neutral. We therefore make no cross-model composed-task claim.

\subsection{Local Interpretations, Not a General Taxonomy}

The evidence supports two useful local descriptions. In the Qwen maths cell,
\texttt{L23H12} is needed mainly at answer positions and behaves as local
reference-answer support; expanded generation controls show that its exact
output damage is not arithmetic-specific. In Llama compare and Mistral dates,
selected prompt-side heads repair prompt processing but their states are inert,
negative, or broad under transfer. These observations motivate answer-side
support and prompt-side maintenance as candidate interpretations in those
cells. They do not establish universal head classes.


\section{Discussion}

The practical lesson is an ordering of claims. Ablation shows that a component
matters for a measured behavior. A linear readout shows that information is
available at its activation. Same-prompt restoration shows that its original
state can repair a particular intervention. None of these establishes that the
component carries a reusable computation. That stronger claim requires moving
the proposed state into a new context and comparing it with controls that
preserve answer identity, computation, or perturbation size.

The controls are not a formality. In Qwen times and Mistral compare, a source
patch moves scores, but matched controls move comparably. Without those
controls, the same observations could be reported as semantic transfer.
Likewise, the compare subspace produces dose-dependent score movement but no
answer switch on adversarial rows. A statistically detectable change is not
automatically a portable computation.

The sensitivity experiments sharpen rather than erase this caution. Known task
state is writable in an underspecified Qwen prompt, and a compact residual state
is overridable in one controlled-model seed. The first result does not test an
explicit competing instruction; the second concerns a residual object in a toy
model. The real-model override behavior is attention-mediated, so neither
positive control licenses reading every failed head patch as absence. The
honest conclusion is that the evidence types dissociate under the tested
interventions.

The local prompt-side and answer-side patterns remain useful. They tell us when
a selected contribution is required and suggest where to look next. They should
not be promoted to a general taxonomy until they replicate across more cells
and survive interventions on MLPs, residual directions, and component
coalitions.


\section{Limitations}
\label{sec:limitations}

\textbf{Assay coverage.} The full matched-control transfer assay covers roughly
8 of 15 model--family cells. The tested cells span all three models and all five
single-step families, but the grid is not exhaustive. Screening breadth must
not be confused with breadth of the complete evidence conjunction.

\textbf{Model scope.} The original evidence matrix uses three instruction-tuned
7--8B models. No larger or reasoning-specialized model is tested. The composed
negative is Qwen-only across two forms: the Llama attempt failed its harness
before injection and is neither a positive nor a negative.

\textbf{Prompt family coverage.} The prompt families are hand-built investigative
probes, not a complete model-native capability ontology. They were chosen for
experimental leverage: anti-correlated subtypes, simple answer formats, and verifiable
correct answers. Results may not transfer to capability classes outside this
set, including open-ended generation, long reasoning trajectories, or
capabilities that are not computation-like. The single-step inventories use 25
items per family and fixed templates; template perturbation and
larger-inventory robustness remain open.

\textbf{Component scope.} The main transfer matrix concerns selected attention
heads. It does not test whether an equivalent state is carried by MLPs,
residual-stream directions, or coalitions of heads. Same-prompt downstream MLP
repair observed elsewhere in the project is not evidence of cross-prompt
portability.

\textbf{Sensitivity scope.} Function-vector recovery validates the write
regime. The controlled model validates the mean-difference transfer operation
for one strong residual-carrier seed. It does not validate the full selection
and representation pipeline, head-level specificity, or sensitivity to the
attention-mediated real-model override object. The failed attention-output
diagnostic is not counted as evidence.

\textbf{Method choices.} The behavioral selection metric, SVD-based
representation audit, and full-state patching choices have not been compared
systematically with alternatives. The results establish a dissociation under
these implementations, not their uniqueness or optimality.

\textbf{No confirmed portable selection state.} No tested attention head
satisfies all four evidence criteria, but absence is not established. A valid
positive would require controlled answer changes under cross-prompt transfer.
Broader negative claims require tests of other components, intervention
granularities, models, prompts, and object classes.


\section{Conclusion}

The selected heads are real causal objects and often contain readable task
information. The central result is that these facts, even when joined with
same-prompt repair, do not establish a reusable computation state. In the
tested attention-head cells, cross-prompt transfer is inert, moves in the wrong
direction, or is matched by controls.

Positive controls show that the intervention can recover known state in a
write regime and a compact residual override state in one controlled-model
seed. They also define the boundary of the claim: the real-model override
negatives are not proven absences, the composed extension is Qwen-only, and
untested components remain open. Mechanistic role claims should therefore name
the exact evidence they rest on. Necessity, decodability, repair, and transfer
are different claims, not interchangeable labels for understanding a
component.


\bibliography{paper}

@article{bair2026cscl,
  title   = {Compressed Sensing for Capability Localization in {LLM}s},
  author  = {Bair, Anna and Xu, Yixuan Even and Sun, Mingjie and Kolter, J. Zico},
  journal = {arXiv preprint arXiv:2603.03335},
  year    = {2026}
}

@article{ahmad2025svd,
  title   = {Beyond Components: Singular Vector-Based Interpretability of Transformer Circuits},
  author  = {Ahmad, Areeb and Joshi, Abhinav and Modi, Ashutosh},
  journal = {arXiv preprint arXiv:2511.20273},
  year    = {2025}
}

@article{qwen2025qwen25,
  title   = {{Qwen2.5} Technical Report},
  author  = {{Qwen Team}},
  journal = {arXiv preprint arXiv:2412.15115},
  year    = {2025}
}

@article{dubey2024llama3,
  title   = {The {Llama} 3 Herd of Models},
  author  = {Dubey, Abhimanyu and Jauhri, Abhinav and Pandey, Abhinav and Kadian, Abhishek and
             Al-Dahle, Ahmad and Letman, Aiesha and Mathur, Akhil and Schelten, Alan and
             Yang, Amy and Fan, Angela and others},
  journal = {arXiv preprint arXiv:2407.21783},
  year    = {2024}
}

@misc{olsson2022induction,
  title     = {In-context Learning and Induction Heads},
  author    = {Olsson, Catherine and Elhage, Nelson and Nanda, Neel and Joseph, Nicholas and
               DasSarma, Nova and Henighan, Tom and Mann, Ben and Askell, Amanda and
               Bai, Yuntao and Chen, Anna and Conerly, Tom and Drain, Dawn and Ganguli, Deep and
               Hatfield-Dodds, Zac and Hernandez, Danny and Johnston, Scott and Jones, Andy and
               Kernion, Jackson and Lovitt, Liane and Ndousse, Kamal and Amodei, Dario and
               Amodei, Daniela and Clark, Jack and Kravec, Shauna and
               Bowman, Sam and Kaplan, Jared and McCandlish, Sam and Olah, Chris},
  howpublished = {Transformer Circuits Thread},
  year      = {2022},
  url       = {https://transformer-circuits.pub/2022/in-context-learning-and-induction-heads/index.html}
}

@inproceedings{wang2022ioi,
  title     = {Interpretability in the Wild: a Circuit for Indirect Object Identification in {GPT-2} Small},
  author    = {Wang, Kevin and Variengien, Alexandre and Conmy, Arthur and Shlegeris, Buck and Steinhardt, Jacob},
  booktitle = {International Conference on Learning Representations},
  year      = {2023},
  url       = {https://arxiv.org/abs/2211.00593}
}

@inproceedings{meng2022rome,
  title   = {Locating and Editing Factual Associations in {GPT}},
  author  = {Meng, Kevin and Bau, David and Andonian, Alex and Belinkov, Yonatan},
  booktitle = {Advances in Neural Information Processing Systems},
  volume  = {35},
  year    = {2022}
}

@article{belinkov2022probing,
  title   = {Probing Classifiers: Promises, Shortcomings, and Advances},
  author  = {Belinkov, Yonatan},
  journal = {Computational Linguistics},
  volume  = {48},
  number  = {1},
  pages   = {207--219},
  year    = {2022}
}

@inproceedings{burns2022ccs,
  title   = {Discovering Latent Knowledge in Language Models Without Supervision},
  author  = {Burns, Collin and Ye, Haotian and Klein, Dan and Steinhardt, Jacob},
  booktitle = {International Conference on Learning Representations},
  year    = {2023}
}

@article{elazar2021amnesic,
  title   = {Amnesic Probing: Behavioral Explanation with Amnesic Counterfactuals},
  author  = {Elazar, Yanai and Ravfogel, Shauli and Jacovi, Alon and Goldberg, Yoav},
  journal = {Transactions of the Association for Computational Linguistics},
  volume  = {9},
  pages   = {160--175},
  year    = {2021},
  url     = {https://arxiv.org/abs/2006.00995}
}

@inproceedings{hewitt2019control,
  title   = {Designing and Interpreting Probes with Control Tasks},
  author  = {Hewitt, John and Liang, Percy},
  booktitle = {Proceedings of the 2019 Conference on Empirical Methods in Natural Language Processing},
  year    = {2019}
}

@inproceedings{vig2020causal,
  title   = {Investigating Gender Bias in Language Models Using Causal Mediation Analysis},
  author  = {Vig, Jesse and Gehrmann, Sebastian and Belinkov, Yonatan and Qian, Sharon and
             Nevo, Daniel and Singer, Yaron and Shieber, Stuart},
  booktitle = {Advances in Neural Information Processing Systems},
  volume  = {33},
  year    = {2020}
}

@inproceedings{conmy2023acdc,
  title   = {Towards Automated Circuit Discovery for Mechanistic Interpretability},
  author  = {Conmy, Arthur and Mavor-Parker, Augustine and Lynch, Aengus and Heimersheim, Stefan and
             Garriga-Alonso, Adri{\`a}},
  booktitle = {Advances in Neural Information Processing Systems},
  volume  = {36},
  year    = {2023}
}

@inproceedings{todd2024functionvectors,
  title   = {Function Vectors in Large Language Models},
  author  = {Todd, Eric and Li, Millicent and Sharma, Arnab Sen and Mueller, Aaron and
             Wallace, Byron C and Bau, David},
  booktitle = {International Conference on Learning Representations},
  year    = {2024}
}

@misc{elhage2022superposition,
  title        = {Toy Models of Superposition},
  author       = {Elhage, Nelson and Hume, Tristan and Olsson, Catherine and Schiefer, Nicholas and
                  Henighan, Tom and Kravec, Shauna and Hatfield-Dodds, Zac and Lasenby, Robert and
                  Drain, Dawn and Chen, Carol and others},
  howpublished = {Transformer Circuits Thread},
  year         = {2022},
  url          = {https://transformer-circuits.pub/2022/toy_model/index.html}
}

@article{yu2024superweight,
  title   = {The Super Weight in Large Language Models},
  author  = {Yu, Mengxia and Wang, De and Shan, Qi and Reed, Colorado J. and Wan, Alvin},
  journal = {arXiv preprint arXiv:2411.07191},
  year    = {2024},
  url     = {https://arxiv.org/abs/2411.07191}
}

@misc{jiang2023mistral,
  title        = {Mistral 7B},
  author       = {Jiang, Albert Q. and Sablayrolles, Alexandre and Mensch, Arthur and
                  Bamford, Chris and Chaplot, Devendra Singh and de las Casas, Diego and
                  Bressand, Florian and Lengyel, Gianna and Lample, Guillaume and
                  Saulnier, Lucile and Renard Lavaud, L{\'e}lio and Lachaux, Marie-Anne and
                  Stock, Pierre and Le Scao, Teven and Lavril, Thibaut and Wang, Thomas and
                  Lacroix, Timoth{\'e}e and El Sayed, William},
  howpublished = {arXiv preprint arXiv:2310.06825},
  year         = {2023},
  url          = {https://arxiv.org/abs/2310.06825}
}

@article{hernandez2023linearconcepts,
  title   = {Linearity of Relation Decoding in Transformer Language Models},
  author  = {Hernandez, Evan and Variengien, Alexandre and Bau, David and Andreas, Jacob},
  journal = {arXiv preprint arXiv:2308.09124},
  year    = {2023},
  url     = {https://arxiv.org/abs/2308.09124}
}

@article{mcdougall2023copysuppression,
  title   = {Copy Suppression: Comprehensively Understanding an Attention Head},
  author  = {McDougall, Callum and Conmy, Arthur and Rushing, Cody and McGrath, Thomas and Nanda, Neel},
  journal = {arXiv preprint arXiv:2310.04625},
  year    = {2023},
  url     = {https://arxiv.org/abs/2310.04625}
}

@inproceedings{makelov2023subspaceillusion,
  title     = {Is This the Subspace You Are Looking for? An Interpretability Illusion for Subspace Activation Patching},
  author    = {Makelov, Aleksandar and Lange, Georg and Nanda, Neel},
  booktitle = {NeurIPS 2023 Workshop on Attributing Model Behavior at Scale},
  year      = {2023},
  url       = {https://arxiv.org/abs/2311.17030}
}

@inproceedings{friedman2024illusions,
  title     = {Interpretability Illusions in the Generalization of Simplified Models},
  author    = {Friedman, Dan and Lampinen, Andrew K. and Dixon, Lucas and Chen, Danqi and Ghandeharioun, Asma},
  booktitle = {Proceedings of the 41st International Conference on Machine Learning},
  series    = {Proceedings of Machine Learning Research},
  volume    = {235},
  pages     = {14035--14059},
  year      = {2024},
  url       = {https://arxiv.org/abs/2312.03656}
}

@article{meloux2025variance,
  title   = {Mechanistic Interpretability as Statistical Estimation: A Variance Analysis},
  author  = {M{\'e}loux, Maxime and Portet, Fran{\c{c}}ois and Peyrard, Maxime},
  journal = {arXiv preprint arXiv:2510.00845},
  year    = {2025},
  url     = {https://arxiv.org/abs/2510.00845}
}

@article{sharkey2025openproblems,
  title   = {Open Problems in Mechanistic Interpretability},
  author  = {Sharkey, Lee and Chughtai, Bilal and Batson, Joshua and Lindsey, Jack and Wu, Jeff and
             Bushnaq, Lucius and Goldowsky-Dill, Nicholas and Heimersheim, Stefan and Ortega, Alejandro and
             Bloom, Joseph and Biderman, Stella and Garriga-Alonso, Adri{\`a} and Conmy, Arthur and
             Nanda, Neel and Rumbelow, Jessica and Wattenberg, Martin and Schoots, Nandi and
             Miller, Joseph and Michaud, Eric J. and Casper, Stephen and Tegmark, Max and
             Saunders, William and Bau, David and Todd, Eric and Geiger, Atticus and Geva, Mor and
             Hoogland, Jesse and Murfet, Daniel and McGrath, Thomas},
  journal = {Transactions on Machine Learning Research},
  year    = {2025},
  url     = {https://arxiv.org/abs/2501.16496}
}

@misc{lan2026auditable,
  title   = {Make Mechanistic Interpretability Auditable: A Call to Develop Guidelines via Continuous Collaborative Reviewing},
  author  = {Lan, Michael and Oozeer, Narmeen Fatimah and Bandi, Chaithanya and Quirke, Philip and
             Meek, Austin and Barez, Fazl and Abdullah, Amirali},
  year    = {2026},
  howpublished = {Preprint, accepted to ICML 2026},
  doi     = {10.5281/zenodo.19671185},
  url     = {https://zenodo.org/records/19671185}
}

@inproceedings{zhang2024patching,
  title     = {Towards Best Practices of Activation Patching in Language Models: 
               Metrics and Methods},
  author    = {Zhang, Fred and Nanda, Neel},
  booktitle = {International Conference on Learning Representations},
  year      = {2024},
  url       = {https://arxiv.org/abs/2309.16042}
}

@inproceedings{hendel2023taskvectors,
  title     = {In-Context Learning Creates Task Vectors},
  author    = {Hendel, Roee and Geva, Mor and Globerson, Amir},
  booktitle = {Findings of the Association for Computational Linguistics: EMNLP 2023},
  pages     = {9318--9333},
  year      = {2023},
  doi       = {10.18653/v1/2023.findings-emnlp.624},
  url       = {https://aclanthology.org/2023.findings-emnlp.624/}
}

@article{heimersheim2024patching,
  title   = {How to Use and Interpret Activation Patching},
  author  = {Heimersheim, Stefan and Nanda, Neel},
  journal = {arXiv preprint arXiv:2404.15255},
  year    = {2024},
  url     = {https://arxiv.org/abs/2404.15255}
}

@inproceedings{yin2025attentionheads,
  title     = {Which Attention Heads Matter for In-Context Learning?},
  author    = {Yin, Kayo and Steinhardt, Jacob},
  booktitle = {Proceedings of the 42nd International Conference on Machine Learning},
  series    = {Proceedings of Machine Learning Research},
  volume    = {267},
  pages     = {72428--72461},
  year      = {2025},
  url       = {https://proceedings.mlr.press/v267/yin25e.html}
}

@article{opielka2026causality,
  title   = {Causality $\neq$ Invariance: Function and Concept Vectors in {LLM}s},
  author  = {Opie{\l}ka, Gustaw and Rosenbusch, Hannes and Stevenson, Claire E.},
  journal = {arXiv preprint arXiv:2602.22424},
  year    = {2026},
  url     = {https://arxiv.org/abs/2602.22424}
}


\appendix

\section{Broader Impacts}
\label{app:broaderimpacts}

This work is foundational interpretability research. Its positive impact is to make
mechanistic evidence standards more precise: behaviorally important, decodable, and
ablation-reversible components should not be promoted to semantic role claims without
matched interventional tests. Better evidence separation can reduce overconfident
claims about model internals and improve safety audits. The main negative impact is
dual-use: sharper localization and intervention methods could also help target model
weaknesses or manipulate behavior. We mitigate this by reporting aggregate
diagnostics on simple computation prompts and a small controlled model rather
than releasing harmful prompts or a deployment method.


\section{Glossary}
\label{app:glossary}

This glossary defines uncommon terms and terms used with a specific operational
meaning in this paper. Linked terms in the main text point here. A definition
states what an assay supports, not a general claim about transformer architecture.

\newcommand{\glossaryentry}[2]{\paragraph{#1}\phantomsection\label{gls:#2}}

\subsection{Evidence Claims}

\glossaryentry{Necessity.}{necessity}
A component is necessary for a measured behavior when removing or zeroing it
reliably damages that behavior. Necessity does not identify what information the
component carries or why the model needs it.

\glossaryentry{Decodability.}{decodability}
Information is decodable when a readout, here a linear nearest-centroid probe,
can predict a task label from an activation. Decodability shows that information
is readable at a site, not that the model uses that readout.

\glossaryentry{Ablation reversal.}{ablation-reversal}
Repair of ablation damage by restoring the component's clean activation in the
same prompt. This establishes local reversibility at the restored token range,
not portability to a different prompt.

\glossaryentry{Cross-prompt transfer.}{cross-prompt-transfer}
Moving an activation from a source prompt into a matched target prompt that asks
for a different computation. A successful transfer must redirect the target
answer toward the source computation more than matched controls do.

\glossaryentry{Interventional generalizability.}{interventional-generalizability}
The property tested by controlled cross-prompt transfer: a proposed internal
state continues to exert its hypothesized causal effect in a new context. The
claim is scoped to the tested component, patch site, prompts, and controls.

\subsection{Assays and Activation Locations}

\glossaryentry{Behavioral ablation screen.}{behavioral-screen}
A search for attention heads whose removal damages one prompt family more than
comparison families. The screen identifies causal targets; it does not assign
those targets a semantic role.

\glossaryentry{Representation audit.}{representation-audit}
An analysis of what information can be read from selected activations. This
paper combines linear probes, singular value decomposition, and surface-form
controls. The audit describes a representation but does not select components.

\glossaryentry{SVD.}{svd}
Singular value decomposition, a matrix factorization used here to identify
dominant directions in activation data. A strong or interpretable SVD direction
is descriptive evidence unless a separate intervention establishes causal use.

\glossaryentry{Activation patching.}{activation-patching}
Replacing an activation in one run with an activation captured from another run.
The interpretation depends on the source, target, patch location, token range,
and behavioral metric.

\glossaryentry{Activation transduction.}{activation-transduction}
The paper's name for cross-prompt activation patching intended to transfer a
requested computation. It is the operation used in the transfer assays, not a
claim that transduction succeeded.

\glossaryentry{Same-prompt restoration.}{same-prompt-restoration}
After ablation, restoring the same prompt's clean activations at selected token
positions. Prompt-only and answer-only restoration localize when a component's
contribution is needed.

\glossaryentry{Prompt-side.}{prompt-side}
A contribution whose ablation damage is substantially repaired by restoring
activations at prompt token positions. This can reflect request processing,
context maintenance, routing, or another pre-generation function; it does not by
itself establish a portable computation-selection state.

\glossaryentry{Answer-side.}{answer-side}
A contribution whose damage is substantially repaired at positions used to
score or generate the answer. In this paper, answer-side is a localization
description rather than a universal head class.

\glossaryentry{Full-trajectory restoration.}{full-trajectory-restoration}
Same-prompt restoration repeated over token ranges such as all prompt positions,
the second half of the prompt, or answer positions. Comparing ranges separates
prompt-side, answer-side, and mixed repair patterns.

\glossaryentry{Recovery fraction.}{recovery-fraction}
The fraction of ablation damage repaired by a restoration intervention. A value
near one means the selected restoration approximately returns the damaged score
to its clean value; zero means no repair. Negative values indicate further harm.

\subsection{Transfer Controls}

\glossaryentry{Matched control.}{matched-control}
A patching condition preserving a potential confound while changing the feature
of interest. A computation-transfer effect must exceed relevant matched controls,
not merely differ from an unpatched baseline.

\glossaryentry{Same-computation control.}{same-computation-control}
Source and target retain the same requested computation while context or surface
form changes. Comparable movement here indicates broad context transfer rather
than a computation switch.

\glossaryentry{Same-answer control.}{same-answer-control}
Source and target share the answer string but require different computations.
Comparable movement here indicates answer-linked or broad state transfer rather
than computation-specific transfer.

\glossaryentry{Same-prompt control.}{same-prompt-control}
The target receives its own clean activation. A near-zero effect checks that the
patching implementation itself does not destabilize the target.

\glossaryentry{Matched-random control.}{matched-random-control}
A random intervention matched to a meaningful patch in scale or norm. It tests
whether output movement follows the proposed content rather than perturbation
magnitude alone.

\glossaryentry{Counterbalanced prompt control.}{counterbalanced-control}
A prompt set designed to separate requested computation from answer string,
format, or template. These controls flag directions dominated by surface
features rather than the intended semantic contrast.

\subsection{Measurements and Scope}

\glossaryentry{Source and target prompt.}{source-target}
The source supplies the activation to be patched. The target receives the patch
and is the prompt the model answers. They are matched on specified surface
features while differing in the computation under test.

\glossaryentry{Model--family cell.}{model-family-cell}
One model paired with one prompt family, for example Qwen with time arithmetic.
Coverage counts in this paper use the model--family cell as their unit.

\glossaryentry{Answer-ordering change.}{answer-ordering-change}
A change in which candidate answer is ranked highest after intervention. The
transfer criterion requires controlled answer reordering; a small log-probability
shift alone shows influence, not a computation switch.

\glossaryentry{Write regime.}{write-regime}
An intervention into an underspecified prompt where the injected state supplies
missing task information. This differs from the harder override regime, where an
intervention must overcome a computation already specified by the target prompt.

\glossaryentry{Object class.}{object-class}
The kind of internal quantity being intervened on, such as an individual head
output, a residual-stream direction, or a coalition of components. Sensitivity
to one object class does not establish sensitivity to another.

\glossaryentry{Passing gate.}{passing-gate}
An assay run whose pre-specified validity checks and positive controls pass, so
its outcome can be interpreted. A failed gate makes a run invalid or
non-diagnostic rather than positive or negative evidence.

\let\glossaryentry\relax

\section{Case Studies}
\label{app:casestudies}

This appendix gives numerical detail for representative cells across all three
models and for the compare subspace. Each subsection states a local
interpretation and then walks through the evidence that supports and limits it.

\subsection{\texttt{qwen maths}: An Answer-Side Logit-Bias Head}

\textbf{Local interpretation:} answer-side reference-answer support; not a
prompt-side portable operation state.

The top-5 set is dominated by rank-1 head \texttt{L23H12}. Activation transduction
using last-4-token activations produces visible logprob movement (source delta $+0.460$,
correct-top 0.969 $\to$ 0.881), but removing \texttt{L23H12} eliminates almost the
entire effect (correct-top unchanged at 0.969). The effect is rank-1 dependent.

\glossaryterm{full-trajectory-restoration}{Full-trajectory restoration}
identifies the role: rank-1 damage (0.259) is almost entirely
answer-restorable (answer-all recovery 1.000, prompt-all recovery 0.022). The head
matters during or after the first answer token, not during prompt processing.

Generation and role-typing audits characterize the role further. Under top-5 ablation,
reference-answer damage is 0.364 while plausible-wrong answers gain probability,
consistent with an answer-side logit-bias head rather than generic continuation damage.
On a 120-prompt arithmetic inventory, rank-1 exact-match damage is 0.092 with 11
margin-crossing rows. However, expanded controls show this is not arithmetic-selective:
control max is 0.100, mostly from \texttt{times}.

\subsection{\texttt{llama digits}: A Mixed Aggregate Where Prompt Recovery Does Not Imply Interventional Generalizability}

\textbf{Role:} mixed aggregate --- answer-side logit-bias heads (ranks 1, 4),
prompt-side stabilizer (rank 3), mixed component (rank 5).

This cell has all the surface ingredients for a portable-state hypothesis: selective
necessity (top-5 selectivity 0.531), a highly subtype-decodable rank-1 head (subtype
accuracy 0.920), and meaningful prompt-second-half restore (recovery 0.422, control
max 0.028).

Activation transduction fails. Patching top-5 prompt-second-half activations gives
source delta $+0.160$ and answer-changed rate 0.013. The source digit property does
not redirect behavior.

Full-trajectory restore explains the aggregate. Ranks 1 and 4 are answer-restorable;
rank 3 is prompt-side (prompt-all recovery 1.003, answer-all recovery $-0.019$);
rank 5 is mixed. Targeted activation transduction on the prompt-side ranks: rank-3
patches give source delta $+0.001$ (inert); rank-5 patches move in the wrong direction.

\subsection{\texttt{qwen times}: A Prompt-Primary Cell With Context-Broad Transfer}

\textbf{Role:} prompt-primary but context-broad. Real prompt-trajectory necessity;
no specific computation-selection state confirmed.

Full-trajectory restore establishes a genuinely prompt-primary locus: top-5 prompt-all
restore recovers 0.722 of zero-all damage and prompt-second-half restores 0.757, while
answer-all restores only 0.478. Rank-1 (\texttt{L0H0}) is strongly prompt-side
(prompt-second-half recovery 0.816, answer-all recovery 0.057).

Activation transduction is control-sensitive. Patching rank-1 prompt-all from a
time-addition source raises source-answer log-probability by 1.209, but
same-computation controls rise more ($+1.338$) and reduce correct-top accuracy
(0.625 $\to$ 0.500). Prompt-all
disambiguation confirms that source answers, source-operation candidates, and impossible
or format distractors all move together, indicating broad source-prompt context transfer
rather than a specific time-operation selection.

\subsection{Llama Comparison: Prompt-Side Repair}

The cleanest available prompt-side targets fail cross-prompt transfer. This
suggests local prompt maintenance and extends the dissociation to a favorable
test case.

Full-trajectory restore classifies the full top-5 as answer-primary (answer-all
recovery 0.760 vs.\ prompt-all recovery 0.354). Rank-level restore reveals a clean
split: ranks 2 and 3 are strongly prompt-side (rank-2: prompt-second-half recovery
0.949, answer-all recovery 0.205; rank-3: prompt-all recovery 1.040, answer-all
recovery $-0.019$), while ranks 1, 4, and 5 are answer-side.

This made ranks 2 and 3 the cleanest available targets for a portable operation
state. Under cross-prompt transfer on
opposite-relation same-candidate rows, the prompt-side ranks are flat or negative:
rank-2 prompt-second-half gives source delta $-0.210$, margin delta $-0.198$;
rank-3 prompt-all gives source delta $-0.015$, margin delta $-0.011$. All-row
movement is small and produces no answer-ordering change. Same-prompt controls confirm
the negative read is not patch instability.

\paragraph{Mistral rows.} The Mistral case studies extend the same role read.
\texttt{mistral dates} rank~1 (\texttt{L25H12}) is cleanly prompt-side
(prompt-all recovery 1.02) but inert under source patching
(source~$\Delta = +0.001$). \texttt{mistral compare} rank~1 (\texttt{L8H1}) is
prompt-primary (prompt-all recovery 0.95) but broad and control-sensitive under
source patching.

\subsection{Soft Computation-Pattern Transfer: The Compare-SV2 Result}

\textbf{Local interpretation:} a decodable ordered-selection neighborhood with
a soft causal bias. It does not constitute a clean interventionally
generalizable computation switch.

Outside the behavioral screen, SVD-based decomposition of Qwen head
\texttt{L14H15} identifies a subspace (\texttt{SV2:SV5})
that carries broader ordered-selection information, activating across numeric
largest/smallest and closest/farthest, digit comparisons, letter ordering, and date
and time relations.

Matched component patching at the local head (\texttt{SV2:SV5} slice only) selectively
lifts source-relation answers on ordered-selection targets while leaving pure surface
controls weaker. Calibrated downstream residual injection at layer~18 amplifies this
signal (source logprob $+0.222$, margin $+0.237$) and scales monotonically with
$\alpha$. However, the direct adversarial test using near-boundary low-margin ordered
rows shows no answer-ordering change and near-zero ordered movement at $\alpha = 0.80$.
Same-answer controls can move comparably to ordered rows.

This result is a decodable ordered-selection neighborhood with a soft causal
bias. Because it produces no answer switch and same-answer controls move
comparably, it does not validate sensitivity to a portable computation state
and does not strengthen the negative head-transfer claim.


\section{Prompt Family Details}
\label{app:prompts}

We use five computation families plus a factual-recall control. Subtype pairs are
anti-correlated (e.g., addition vs.\ subtraction) so that selecting one computation
should suppress the other, making the families natural probes for polysemantically
packed representations \citep{elhage2022superposition}. Table~\ref{tab:prompt-families}
summarises the families.

\begin{table}[h]
\centering
\caption{Prompt families. Subtype pairs are anti-correlated and may share
representational space. Each family uses a fixed template with candidate options
and a reference answer; the model is scored on reference-answer log-probability.
All families have 25 items.}
\label{tab:prompt-families}
\small
\setlength{\tabcolsep}{4pt}
\begin{tabular}{@{}>{\raggedright\arraybackslash}p{0.13\textwidth}>{\raggedright\arraybackslash}p{0.32\textwidth}>{\raggedright\arraybackslash}p{0.46\textwidth}@{}}
\toprule
Family & Subtypes & Template style \\
\midrule
\texttt{maths}   & add, subtract, multiply, divide
                 & ``What is $a \oplus b$?'' \\
\texttt{compare} & largest, smallest, middle, closest, farthest
                 & ``Which number is the largest/smallest\ldots?'' (4 candidates) \\
\texttt{digits}  & odd, even, prime, composite
                 & ``Is $n$ odd or even / prime or composite?'' \\
\texttt{dates}   & before, after
                 & ``Which date comes before/after\ldots?'' (2 dates) \\
\texttt{times}   & time-add, time-subtract
                 & ``What time is it $t$ hours after/before $T$?'' \\
\texttt{facts}   & factual recall (control)
                 & ``What is the capital of\ldots?'' \\
\bottomrule
\end{tabular}
\end{table}

Counterbalanced control construction for SVD experiments: each semantic prompt
(subtype $A$) is paired with a surface control prompt that shares the answer
string but uses a different subtype (e.g., the answer ``9'' appearing in both
an addition and a multiplication prompt). Controls are drawn from the same
template with candidate sets chosen so the reference answer is identical.
This isolates semantic (subtype-specific) SVD directions from surface
(answer-string-related) directions.

For activation transduction, pairs are constructed by matching template,
candidate set, and answer format across opposite-subtype prompts
(e.g., source: ``which is largest?'', target: ``which is smallest?''),
sharing the same candidate numbers but requiring the opposite relation.
Same-answer controls replace the source with a prompt from a different subtype
that has the same answer string.

\section{Behavioral Screening Details}
\label{app:css}

For each model-family cell, selectivity is computed as:
\[
s_h = \Delta_{\text{target}}(h) - \max_{f \neq \text{target}} \Delta_f(h)
\]
where $\Delta_f(h)$ is the reference-answer log-probability change for family $f$ under
zero-ablation of head $h$. Heads are ranked by $s_h$ descending, and the top-$k$
cumulative set greedily adds heads in that order.

Random-mask recovery runs use: (main pass) 256 stratified masks with 32 heads zeroed
per mask, OMP recovery with 20 nonzero coefficients; (follow-up diagnostic) 1024 masks
with 8 heads per mask.

Subset-lattice evaluation measures all $2^k - 1$ nonempty subsets of the top-5 set.
Reported metrics per subset: target damage, maximum control damage, selectivity, and
per-head marginal contribution.

Table~\ref{tab:css-heads} lists the Qwen, Llama and Mistral selected top-5 sets.
Selectivity is target-family damage minus maximum control-family damage. Heads are
listed in rank order (descending selectivity within the cumulative top-5 screen).

\begin{table}[t]
\centering
\caption{Behavioral-screen top-5 selectivity summary. Target damage is reference-answer logprob
change under top-5 ablation. Selectivity is target damage minus control maximum.
13 of 18 cells are cleanly selective; 2 are selective with
broader collateral damage; the 3 remaining cells fail selectivity.}
\label{tab:css-selectivity}
\setlength{\tabcolsep}{3pt}
\begin{tabular}{llrrrr}
\toprule
Model & Family & Target Damage & Control Max & Selectivity & Read \\
\midrule
\texttt{llama} & \texttt{maths}   & 0.166 & 0.005    & 0.161 & clean \\
\texttt{llama} & \texttt{facts}   & 0.271 & $-$0.007 & 0.278 & clean \\
\texttt{llama} & \texttt{dates}   & 1.206 & 0.721    & 0.485 & clean \\
\texttt{llama} & \texttt{times}   & 0.763 & 0.454    & 0.309 & clean \\
\texttt{llama} & \texttt{digits}  & 0.545 & 0.014    & 0.531 & clean \\
\texttt{llama} & \texttt{compare} & 0.611 & 0.098    & 0.513 & clean \\
\texttt{qwen} & \texttt{maths}   & 0.364 & 0.051    & 0.313 & clean \\
\texttt{qwen} & \texttt{times}   & 1.875 & 0.486    & 1.389 & clean \\
\texttt{qwen} & \texttt{digits}  & 0.242 & 0.038    & 0.204 & clean \\
\texttt{qwen} & \texttt{facts}   & 0.403 & 0.282    & 0.122 & broad \\
\texttt{qwen} & \texttt{dates}   & large & large    & low   & not selective \\
\texttt{qwen} & \texttt{compare} & large & large    & low   & not selective \\
\texttt{mistral} & \texttt{dates}   & 0.901 & 0.068    & 0.833 & clean \\
\texttt{mistral} & \texttt{compare} & 0.639 & 0.073    & 0.566 & clean \\
\texttt{mistral} & \texttt{times}   & 0.499 & 0.019    & 0.480 & clean \\
\texttt{mistral} & \texttt{facts}   & 0.288 & $-$0.028 & 0.316 & clean \\
\texttt{mistral} & \texttt{digits}  & 0.543 & 0.240    & 0.303 & broad \\
\texttt{mistral} & \texttt{maths}   & 0.951 & 0.953    & $-$0.002 & not selective \\
\bottomrule
\end{tabular}
\end{table}

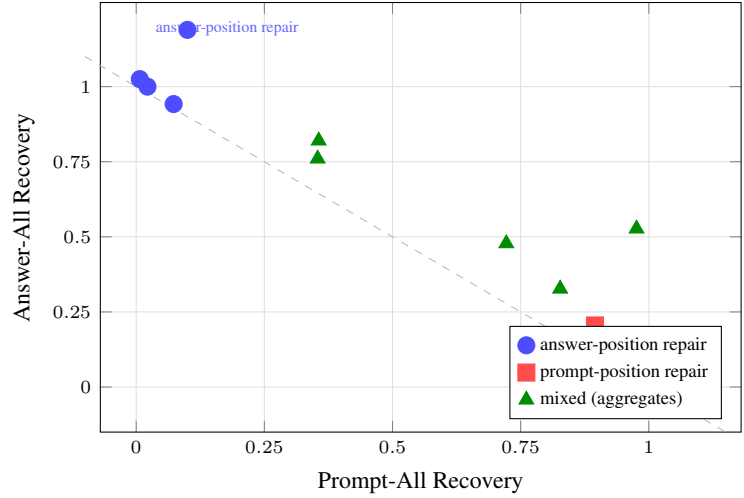
\begin{figure}[t]
\centering
\begin{tikzpicture}
\begin{axis}[
  width=0.72\linewidth, height=0.52\linewidth,
  xlabel={Prompt-All Recovery},
  ylabel={Answer-All Recovery},
  xmin=-0.07, xmax=1.18,
  ymin=-0.15, ymax=1.28,
  xtick={0, 0.25, 0.5, 0.75, 1.0},
  ytick={0, 0.25, 0.5, 0.75, 1.0},
  grid=both,
  grid style={line width=0.3pt, draw=gray!25},
  legend pos=south east,
  legend style={font=\scriptsize, cells={anchor=west}},
  tick label style={font=\scriptsize},
  label style={font=\small},
  clip=false,
]
\addplot[dashed, gray!55, domain=-0.1:1.15, samples=2, forget plot] {1 - x};

\addplot[only marks, mark=*, mark size=3.2pt, blue!70]
  coordinates {
    (0.022, 1.000)
    (0.073, 0.942)
    (0.007, 1.025)
    (0.100, 1.189)
  };
\addlegendentry{answer-position repair}

\addplot[only marks, mark=square*, mark size=3.2pt, red!70]
  coordinates {
    (1.003, -0.019)
    (0.895,  0.205)
    (1.040, -0.019)
    (0.816,  0.057)
    (1.016, -0.018)
    (0.948,  0.004)
  };
\addlegendentry{prompt-position repair}

\addplot[only marks, mark=triangle*, mark size=3.2pt, green!55!black]
  coordinates {
    (0.356, 0.820)
    (0.354, 0.760)
    (0.722, 0.478)
    (0.976, 0.527)
    (0.827, 0.327)
  };
\addlegendentry{mixed (aggregates)}

\node[font=\tiny, blue!60, anchor=north west] at (axis cs: 0.02, 1.25)
  {answer-position repair};
\node[font=\tiny, red!60, anchor=south east]  at (axis cs: 1.10, 0.05)
  {prompt-position repair};
\end{axis}
\end{tikzpicture}
\caption{Same-prompt restoration: prompt-all vs.\ answer-all recovery for the main
tested heads and aggregates. Answer-position repair (blue circles) clusters in the upper-left;
prompt-position repair (red squares) clusters in the lower-right. The dashed diagonal
($\text{prompt-all} + \text{answer-all} = 1$)
separates the two timing patterns. Aggregate top-$k$ sets (green triangles)
fall between clusters, confirming they mix contributions across constituent ranks.}
\label{fig:scatter}
\end{figure}

\begin{table*}[t]
\centering
\caption{Selected top-5 head sets per family, with models shown left to right. Rank order
is by selectivity within each model's cumulative screen. Sel.\ = selectivity
(target damage $-$ control max).}
\label{tab:css-heads}
\setlength{\tabcolsep}{2.2pt}
\resizebox{\textwidth}{!}{%
\begin{tabular}{llrrr rrr rrr}
\toprule
Family & Rank
& \multicolumn{3}{c}{\texttt{qwen}}
& \multicolumn{3}{c}{\texttt{llama}}
& \multicolumn{3}{c}{\texttt{mistral}} \\
\cmidrule(lr){3-5}\cmidrule(lr){6-8}\cmidrule(lr){9-11}
& & Head & Target & Sel. & Head & Target & Sel. & Head & Target & Sel. \\
\midrule
\texttt{maths} & 1 & \texttt{L23H12} & 0.258 & 0.170 & \texttt{L22H28} & 0.042 & 0.040 & \texttt{L15H3}  & 0.039 & 0.030 \\
& 2 & \texttt{L0H4}   & 0.041 & 0.038 & \texttt{L18H18} & 0.032 & 0.030 & \texttt{L14H26} & 0.086 & 0.012 \\
& 3 & \texttt{L22H13} & 0.049 & 0.017 & \texttt{L0H23}  & 0.032 & 0.027 & \texttt{L0H29}  & 0.877 & 0.001 \\
& 4 & \texttt{L23H16} & 0.024 & 0.024 & \texttt{L28H17} & 0.040 & 0.017 & \texttt{L10H23} & 0.025 & 0.022 \\
& 5 & \texttt{L4H17}  & 0.013 & 0.013 & \texttt{L14H3}  & 0.038 & 0.017 & \texttt{L14H13} & 0.035 & 0.012 \\
\addlinespace
\texttt{times} & 1 & \texttt{L0H0}  & 1.088 & 0.627 & \texttt{L16H21} & 0.738 & 0.518 & \texttt{L17H25} & 0.170 & 0.171 \\
& 2 & \texttt{L21H2} & 0.304 & 0.297 & \texttt{L14H18} & 0.091 & 0.049 & \texttt{L4H2}   & 0.132 & 0.132 \\
& 3 & \texttt{L17H9} & 0.174 & 0.155 & \texttt{L13H8}  & 0.057 & 0.047 & \texttt{L21H10} & 0.132 & 0.098 \\
& 4 & \texttt{L3H3}  & 0.077 & 0.060 & \texttt{L8H2}   & 0.047 & 0.052 & \texttt{L1H3}   & 0.108 & 0.090 \\
& 5 & \texttt{L4H12} & 0.124 & 0.029 & \texttt{L11H16} & 0.061 & 0.039 & \texttt{L0H31}  & 0.100 & 0.075 \\
\addlinespace
\texttt{digits} & 1 & \texttt{L22H1}  & 0.154 & 0.150 & \texttt{L28H19} & 0.177 & 0.170 & \texttt{L15H0}  & 0.150 & 0.108 \\
& 2 & \texttt{L23H13} & 0.071 & 0.068 & \texttt{L18H4}  & 0.137 & 0.137 & \texttt{L20H29} & 0.162 & 0.098 \\
& 3 & \texttt{L19H19} & 0.033 & 0.001 & \texttt{L0H5}   & 0.119 & 0.100 & \texttt{L15H29} & 0.107 & 0.103 \\
& 4 & \texttt{L22H2}  & 0.007 & 0.004 & \texttt{L14H12} & 0.103 & 0.104 & \texttt{L20H31} & 0.124 & 0.085 \\
& 5 & \texttt{L22H10} & 0.021 & 0.001 & \texttt{L11H23} & 0.095 & 0.068 & \texttt{L0H27}  & 0.175 & 0.058 \\
\addlinespace
\texttt{compare} & 1 & \texttt{L0H24} & 2.286 & 0.280 & \texttt{L14H4}  & 0.126 & 0.125 & \texttt{L8H1}   & 0.175 & 0.057 \\
& 2 & \texttt{L0H26} & 2.575 & 0.099 & \texttt{L8H11}  & 0.122 & 0.110 & \texttt{L12H6}  & 0.082 & 0.075 \\
& 3 & \texttt{L0H6}  & 0.303 & 0.065 & \texttt{L6H4}   & 0.123 & 0.104 & \texttt{L11H13} & 0.089 & 0.062 \\
& 4 & \texttt{L13H15}& 0.126 & 0.097 & \texttt{L16H30} & 0.167 & 0.074 & \texttt{L13H30} & 0.066 & 0.073 \\
& 5 & \texttt{L0H15} & 0.108 & 0.109 & \texttt{L24H20} & 0.091 & 0.084 & \texttt{L11H5}  & 0.068 & 0.066 \\
\addlinespace
\texttt{dates} & 1 & \texttt{L22H12} & 0.360 & 0.347 & \texttt{L25H30} & 0.416 & 0.381 & \texttt{L25H12} & 0.456 & 0.456 \\
& 2 & \texttt{L0H3}   & 3.290 & 0.037 & \texttt{L25H29} & 0.348 & 0.347 & \texttt{L7H18}  & 0.346 & 0.256 \\
& 3 & \texttt{L0H22}  & 0.471 & 0.154 & \texttt{L21H27} & 0.256 & 0.251 & \texttt{L30H19} & 0.262 & 0.259 \\
& 4 & \texttt{L1H3}   & 0.139 & 0.103 & \texttt{L5H22}  & 0.332 & 0.176 & \texttt{L25H13} & 0.252 & 0.243 \\
& 5 & \texttt{L1H9}   & 0.164 & 0.075 & \texttt{L15H13} & 0.787 & 0.064 & \texttt{L20H28} & 0.288 & 0.203 \\
\addlinespace
\texttt{facts} & 1 & \texttt{L0H25} & 7.293 & 5.273 & \texttt{L31H0}  & 0.080 & 0.076 & \texttt{L2H11}  & 0.156 & 0.157 \\
& 2 & \texttt{L0H1}  & 4.726 & 1.514 & \texttt{L9H6}   & 0.054 & 0.053 & \texttt{L3H28}  & 0.111 & 0.094 \\
& 3 & \texttt{L0H23} & 0.403 & 0.064 & \texttt{L6H22}  & 0.052 & 0.051 & \texttt{L31H25} & 0.101 & 0.099 \\
& 4 & \texttt{L0H27} & 0.314 & 0.057 & \texttt{L31H1}  & 0.055 & 0.045 & \texttt{L2H21}  & 0.085 & 0.101 \\
& 5 & \texttt{L14H20}& 0.145 & 0.119 & \texttt{L27H28} & 0.050 & 0.044 & \texttt{L3H31}  & 0.079 & 0.061 \\
\bottomrule
\end{tabular}%
}
\end{table*}

\section{SVD Analysis Details}
\label{app:svd}

SVD decomposition of head residual contribution matrices exposes interpretable
directions. The $\eta^2$ statistic measures the fraction of variance in the
first singular value accounted for by the requested-computation label, relative to
a shuffle null. The SVD excess is the ratio of the first singular value to the median;
values substantially above 1 indicate a dominant structured direction.

Counterbalanced controls pair each semantic target prompt with a surface control
matching the answer string but differing in requested computation. An SVD direction
that shifts comparably under surface controls is classified as nuisance-dominant.

Selected SVD results for the main case study heads are in Table~\ref{tab:svd-summary}.
Family-level probe accuracy (top-5 concatenated residuals, nearest-centroid) ranges
from 0.76 to 0.93 across models and prompt families.

\begin{table}[h]
\centering
\caption{SVD-based representation audit summary. Probe accuracy is family-level
nearest-centroid classification on top-5 concatenated residuals. Subtype accuracy
is for the listed head where reported. A direction is flagged nuisance-dominant if
it shifts comparably under surface-control prompts.}
\label{tab:svd-summary}
\small
\setlength{\tabcolsep}{4pt}
\resizebox{\textwidth}{!}{%
\begin{tabular}{lllrrl}
\toprule
Model & Family & Head & Probe acc. & Subtype acc. & SVD note \\
\midrule
\texttt{qwen}    & \texttt{maths}   & \texttt{L23H12} & ---   & ---   & rank-1 dominant; answer-side \\
\texttt{qwen}    & \texttt{times}   & \texttt{L0H0}   & ---   & ---   & prompt-side; broad context \\
\texttt{qwen}    & \texttt{compare} & \texttt{L14H15} & ---   & ---   & outside-screen SV2:SV5 ordered-selection subspace \\
\texttt{llama}   & \texttt{digits}  & \texttt{L28H19} & ---   & 0.920 & subtype-decodable; answer-side \\
\texttt{llama}   & \texttt{compare} & \texttt{L14H4}  & ---   & ---   & answer-primary; prompt-side in R2,R3 \\
\texttt{mistral} & \texttt{dates}   & \texttt{L25H12} & 0.933 & 0.560 & prompt-side; strong SVD alignment; transduction inert \\
\texttt{mistral} & \texttt{compare} & \texttt{L8H1}   & 0.913 & 0.040 & family-readable; weak subtype chart; broad transduction \\
\texttt{mistral} & \texttt{digits}  & \texttt{L15H0}  & 0.887 & 0.840 & strong rank-1 subtype readout \\
\texttt{mistral} & \texttt{times}   & \texttt{L17H25} & 0.760 & 0.600 & family-readable; mixed subtype readout \\
\bottomrule
\end{tabular}
}
\end{table}

\section{Activation Transduction Protocol}
\label{app:transduction}

Source and target prompts in each transduction pair are matched on: surface template,
candidate item set, answer format, and target token position. They differ in the
requested computation (e.g., source: largest, target: smallest). Both source-to-target
and target-to-source directions are tested where applicable.

Patch modes tested per cell:
\begin{itemize}
\item Full head state at specified token position range.
\item SVD component slice (\texttt{SV2:SV5} for compare-SV2 experiments).
\item Rank-specific prompt-second-half, prompt-all patches.
\item Residual-delta injection at downstream layers (compare-SV2 downstream experiments).
\end{itemize}

Same-answer control construction: for each (source, target) pair, replace the source
with a prompt that has the same answer but a different requested computation.
Same-computation controls replace the source with a different surface form of the same
computation.

Alpha stress tests for the compare-SV2 downstream residual injection: the causal
handle scales monotonically with injection magnitude. At $\alpha = 0.20$: source
logprob $+0.067$, margin $+0.072$. At $\alpha = 0.40$: source $+0.127$, margin
$+0.145$. At $\alpha = 0.80$: source $+0.222$, margin $+0.237$ on the calibrated
held-out set. However, on the adversarial low-margin ordered rows ($n=10$ rows
within margin 1.25, 6 source-best-wrong), the $\alpha = 0.80$ patch produces
answer-changed rate 0.000 and near-zero ordered movement, indicating the soft causal
handle does not flip decisions on genuinely marginal prompts.

Hard-row inventory for exact-output audits: for \texttt{qwen maths}, 120 prompts
were scored under top-5 ablation. Exact-match damage is 0.092 (11 rows), with
control-max damage 0.100 (mostly \texttt{times}), confirming the effect is not
maths-selective. For \texttt{llama digits}, exact-match damage under top-5
ablation is target-skewed (0.075 target vs.\ 0.025 control maximum).

Exact-output audit methodology: each prompt is run under greedy decoding with and
without ablation. A ``damaged'' row is one where the greedy first token changes from
correct to a different candidate under ablation. Margin is defined as the logprob
difference between the correct answer token and the best competing candidate token.

\section{Compute Resources}
\label{app:compute}

All experiments ran on Google Cloud Platform (GCP) virtual machines. Two machine types
were used:

\begin{itemize}
\item \textbf{g2-standard-8}: 1$\times$ NVIDIA L4 GPU (24\,GB VRAM), 8 vCPUs, 32\,GB
RAM. Used for smaller experiments: representation audits, generation audits, and
single-head analysis scripts.
\item \textbf{a2-highgpu-1g}: 1$\times$ NVIDIA A100 GPU (40\,GB VRAM), 12 vCPUs, 85\,GB
RAM. Used for restoration, activation transfer, random-mask recovery,
and subset lattice experiments.
\end{itemize}

Approximate per-experiment runtimes (including model load):

\begin{itemize}
\item Behavioral screen (one model, all families): $\approx$\,2--3\,hours on A100.
\item Full-trajectory restore (one model + family): $\approx$\,22--60\,minutes on A100.
\item Activation transduction (one model + family, all patch targets): $\approx$\,1--2\,hours on A100.
\item Subset-lattice evaluation (one model + family): $\approx$\,45--90\,minutes on A100.
\item Representation audit (one full-coverage model, all families): $\approx$\,1\,hour on A100.
\item Random-mask recovery (256 masks, one cell): $\approx$\,30--60\,minutes on A100.
\item SVD and downstream patching experiments: $\approx$\,30--90\,minutes each on A100.
\item Controlled-model sensitivity experiment: three tiny-model training seeds
and interventions completed in minutes on one L4 VM.
\end{itemize}

Across all experiments reported in the paper, the estimated total compute is
approximately 60--80 A100-GPU-hours plus the small L4 controlled-model run. All
large-model experiments are inference-time forward passes on publicly released
checkpoints; only the two-layer controlled model is trained from scratch.

\section{Extended Numerical Results}
\label{app:numerical}

\begin{table}[h]
\centering
\caption{Same-prompt restoration results for main case studies. Recovery fraction $=$
(zero-all damage $-$ remaining damage after restoration) / zero-all damage. Values near 1.0 indicate full
recovery at that position range; near 0 indicate no recovery.}
\label{tab:restore}
\scriptsize
\setlength{\tabcolsep}{3pt}
\resizebox{\textwidth}{!}{%
\begin{tabular}{llrrrl}
\toprule
Cell & Head / Rank & Zero-All & Prompt-All & Answer-All & Role \\
\midrule
\texttt{qwen maths}       & rank 1 \texttt{L23H12} & 0.259 & 0.022    & 1.000    & answer-side \\
\texttt{qwen maths}       & top-5                  & 0.364 & 0.073    & 0.942    & answer-side \\
\texttt{llama digits}  & rank 1 \texttt{L28H19} & 0.177 & 0.007    & 1.025\,$\dagger$ & answer-side \\
\texttt{llama digits}  & rank 3 \texttt{L0H5}   & 0.118 & 1.003    & $-$0.019 & prompt-side \\
\texttt{llama digits}  & rank 4 \texttt{L14H12} & 0.103 & 0.100    & 1.189    & answer-side \\
\texttt{llama digits}  & rank 5 \texttt{L11H23} & 0.096 & 0.743\,$\ddagger$ & ---      & mixed \\
\texttt{llama digits}  & top-5                  & 0.545 & 0.356    & 0.820    & answer-primary mixed \\
\texttt{qwen times}       & rank 1 \texttt{L0H0}   & ---   & 0.816\,$\ddagger$ & 0.057    & prompt-side \\
\texttt{qwen times}       & rank 2 \texttt{L21H2}  & ---   & ---      & 1.001    & answer-side \\
\texttt{qwen times}       & top-5                  & 1.875 & 0.722    & 0.478    & prompt-primary \\
\texttt{llama compare} & rank 2 \texttt{L8H11}  & ---   & 0.895    & 0.205    & prompt-side \\
\texttt{llama compare} & rank 3 \texttt{L6H4}   & ---   & 1.040    & $-$0.019 & prompt-side \\
\texttt{llama compare} & top-5                  & 0.611 & 0.354    & 0.760    & answer-primary \\
\texttt{mistral dates}  & rank 1 \texttt{L25H12} & 0.453 & 1.016\,$\ddagger$ & $-$0.018 & prompt-side \\
\texttt{mistral compare} & rank 1 \texttt{L8H1}  & 0.177 & 0.948\,$\ddagger$ & 0.004    & prompt-side \\
\texttt{mistral compare} & top-5                 & 0.642 & 0.827\,$\ddagger$ & 0.327    & prompt-primary mixed \\
\bottomrule
\end{tabular}
}

\smallskip
\footnotesize
$\dagger$ best restore is answer-rest (1.025). \quad $\ddagger$ best restore is prompt-second-half.
\end{table}

\begin{table}[h]
\centering
\caption{Activation transduction results. Source $\Delta$ $=$ source-answer logprob
change. Margin $\Delta$ $=$ source-vs.-correct margin change. Answer Changed $=$ fraction
of prompts where top candidate changes. Controls show the same metric for the most
relevant control condition.}
\label{tab:transduction}
\scriptsize
\setlength{\tabcolsep}{3pt}
\resizebox{\textwidth}{!}{%
\begin{tabular}{llrrrrl}
\toprule
Cell & Patch target & Src $\Delta$ & Margin $\Delta$ & Correct Top & Ans.\ Ch. & Controls \\
\midrule
\texttt{qwen maths}       & top-5 last-4             & $+0.460$ & $+0.608$ & 0.969$\to$0.881 & --- & rank-1 dependent \\
\texttt{llama digits}  & top-5 prompt-2nd-half    & $+0.160$ & $+0.183$ & 0.850$\to$0.838 & 0.013 & same-property $\approx$0 \\
\texttt{llama digits}  & rank 3 prompt-2nd-half   & $+0.001$ & $-$0.001 & 0.850$\to$0.850 & 0.009 & inert \\
\texttt{llama digits}  & rank 4 prompt-2nd-half   & $+0.093$ & $+0.105$ & 0.850$\to$0.850 & 0.003 & weak \\
\texttt{qwen times}       & rank 1 prompt-all        & $+1.209$ & $+1.489$ & --- & --- & same-comp: $+1.338$ \\
\texttt{qwen times}       & top-5 prompt-all         & $+1.328$ & $+1.538$ & 0.625 & --- & same-comp: $+1.600$ \\
\texttt{llama compare} & rank 2 prompt-2nd-half   & $-$0.210 & $-$0.198 & --- & 0.000 & same-prompt $\approx$0 \\
\texttt{llama compare} & rank 3 prompt-all        & $-$0.015 & $-$0.011 & --- & 0.000 & same-prompt $\approx$0 \\
\texttt{llama compare} & rank 2+3 prompt-2nd-half & $-$0.211 & $-$0.196 & --- & 0.000 & same-prompt $\approx$0 \\
\texttt{mistral compare} & rank 1 prompt-all       & $+0.126$ & $+0.236$ & 0.730$\to$0.697 & 0.046 & same-rel: $+0.279$ margin \\
\texttt{mistral compare} & top-5 prompt-all        & $+0.088$ & $+0.286$ & 0.730$\to$0.645 & 0.099 & same-answer: 0.725$\to$0.600 \\
\texttt{mistral dates}   & rank 1 prompt-all       & $+0.001$ & $+0.001$ & 1.000$\to$1.000 & 0.000 & inert \\
\texttt{mistral dates}   & top-5 prompt-all        & $-$0.006 & $-$0.020 & 1.000$\to$1.000 & 0.000 & source-opposite \\
\bottomrule
\end{tabular}
}
\end{table}

\begin{table}[h]
\centering
\caption{Generation audit summary. EM damage $=$ fraction of prompts where greedy
completion changes from correct to incorrect under ablation.}
\label{tab:generation}
\scriptsize
\setlength{\tabcolsep}{3pt}
\resizebox{\textwidth}{!}{%
\begin{tabular}{llrrrr}
\toprule
Cell & Head set & EM Damage & Control Max & EM Rows & Read \\
\midrule
\texttt{qwen maths} & rank 1 \texttt{L23H12}  & 0.092    & 0.100 & 11/120 & not maths-selective \\
\texttt{qwen maths} & top-5                   & 0.100    & 0.175 & 13/120 & not maths-selective \\
\texttt{qwen maths} & top-5 without rank 1    & $-$0.008 & ---   & 0/120  & null \\
\texttt{llama digits} & top-5              & 0.075    & 0.025 & ---    & target-skewed \\
\texttt{llama digits} & top-5 without rank 5 & 0.042  & 0.025 & ---    & target-skewed \\
\bottomrule
\end{tabular}
}
\end{table}

\section{Skeptical Analysis of the Behavioral Screen}
\label{app:skeptical-css}

We compare selected top-5 sets with random additive head sets. The selected sets
are statistically extreme in all audited cells, so the screen is not choosing
arbitrary heads. This does not imply that every cell is sharply sparse. Qwen
digits, maths, and times have the strongest concentration into low effective
head counts. Several Llama and Mistral cells are non-random but broad-tailed,
with many mildly positive heads. Exact top-5 reuse across families is zero in
all three models; the data do not support a simple shared-head account of all
computation-like prompts.

\section{Skeptical Analysis of SVD}
\label{app:skeptical-svd}

The representation structure is also real but not uniquely mechanistic:
441/468 audited alignment rows exceed the 95th percentile of a shuffled-label
null, and many selected states are strongly readable. A layer-matched random
head atlas nevertheless matches or exceeds the selected-head atlas on family
separability and SVD concentration. Low-rank structure therefore supports a
descriptive representation claim, not by itself a unique causal-component or
polysemantic-packing claim.


\end{document}